\documentclass[11pt, a4paper, singlecolumn]{article}

\usepackage{color}

\usepackage[utf8]{inputenc}
\usepackage[english]{babel}

\usepackage[top=2cm, bottom=2cm, left=1.9cm, right=1.9cm]{geometry}

\usepackage[title]{appendix}

\usepackage{titlesec}
\titlespacing{\section}{0pt}{\parskip}{\parskip}
\titlespacing{\subsection}{0pt}{\parskip}{\parskip}
\titlespacing{\subsubsection}{0pt}{\parskip}{\parskip}

\usepackage{indentfirst}
\usepackage[export]{adjustbox}
\usepackage{graphicx}
\usepackage{amssymb}
\usepackage{amsmath}

\usepackage[dvipsnames]{xcolor}

\usepackage[figurename=Fig.]{caption}
\usepackage{caption}
\usepackage[colorlinks, citecolor=ForestGreen]{hyperref}
\usepackage{color, soul}
\usepackage{cite}

\usepackage{abstract}
\usepackage{multirow}
\usepackage[normalem]{ulem}
\usepackage{array}
\newcolumntype{*}{>{\global\let\currentrowstyle\relax}}
\newcolumntype{^}{>{\currentrowstyle}}

\title{\textbf{Development of Automatic Tree Counting Software from UAV Based Aerial Images With Machine Learning}\footnote{This study was supported by Siirt University Scientific Research Projects Coordinatorship (SIÜBAP) within the scope of project numbered 2017-SIÜFEB-21.}}
\date{\vspace*{2pt}}
\author{
	\normalsize \textbf{Musa Ataş}\\
	\normalsize Siirt University\\
	\normalsize Computer Engineering Department\\
	\normalsize musa.atas@siirt.edu.tr
	\and
	\normalsize \textbf{Ayhan Talay}\\
	\normalsize Siirt University\\
	\normalsize Computer Engineering Department\\
	\normalsize atalay65@siirt.edu.tr
}

\begin{document}
\maketitle
%%---------------------------------
%% edit by 
%\thispagestyle{fancy}
%%---------------------------------

\begin{abstract}{
	\vspace*{-1.5em}
	\it Unmanned aerial vehicles (UAV) are used successfully in many application areas such as military, security, monitoring, emergency aid, tourism, agriculture, and forestry. This study aims to automatically count trees in designated areas on the Siirt University campus from high-resolution images obtained by UAV. Images obtained at 30 meters height with 20\% overlap were stitched offline at the ground station using Adobe Photoshop's photo merge tool. The resulting image was denoised and smoothed by applying the 3x3 median and mean filter, respectively. After generating the orthophoto map of the aerial images captured by the UAV in certain regions, the bounding boxes of different objects on these maps were labeled in the modalities of HSV (Hue Saturation Value), RGB (Red Green Blue) and Gray.  Training, validation, and test datasets were generated and then have been evaluated for classification success rates related to tree detection using various machine learning algorithms. In the last step, a ground truth model was established by obtaining the actual tree numbers, and then the prediction performance was calculated by comparing the reference ground truth data with the proposed model. It is considered that significant success has been achieved for tree count with an average accuracy rate of 87\% obtained using the MLP classifier in predetermined regions.
	
}\end{abstract}


\section{Introduction}

\begin{figure}[!h]
	\centering
	\includegraphics[width=0.8\textwidth]{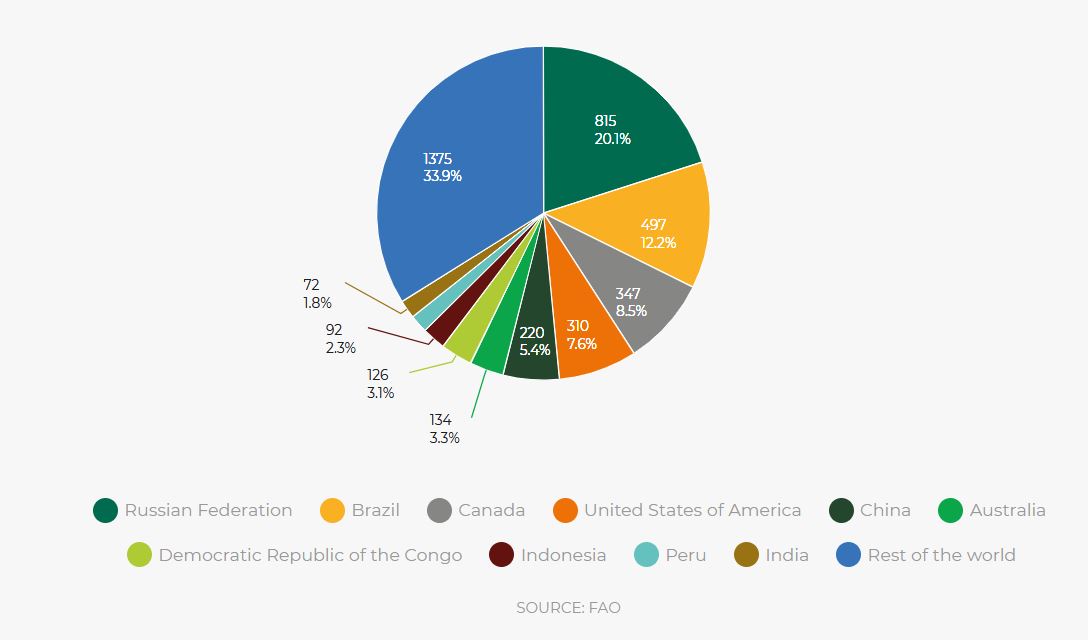}
	\caption{Forest sizes on the basis of countries.}
	\label{fig:world_forest}
\end{figure}

\begin{figure}[!h]
	\centering
	\includegraphics[width=0.5\textwidth]{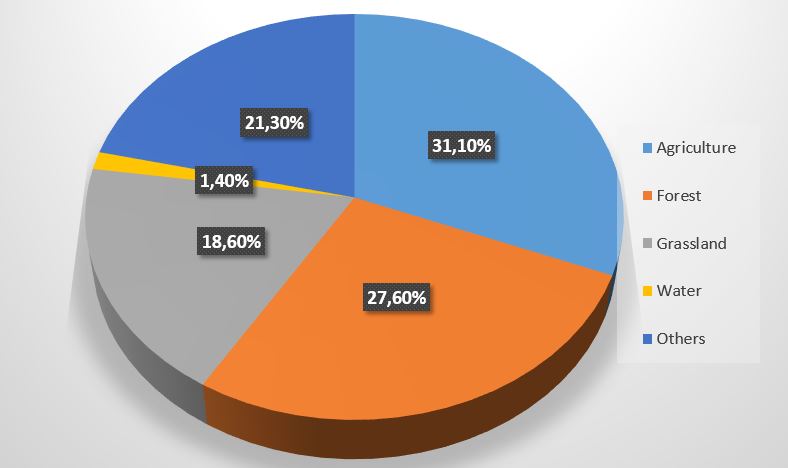}
	\caption{Turkey's land usage.}
	\label{fig:Turkey_forest}
\end{figure}

Since forest ecosystems retain more biodiversity than other ecosystems, they are a very important component for the continuity of life. Forests cover about 31 percent of the world's land area. As seen in Fig. \ref{fig:world_forest}, more than half of the world's forests are found in five countries including Russia, Brazil, Canada, the USA, and China \cite{fao}. Similarly, according to the data in Fig. \ref{fig:Turkey_forest}, approximately 27.6\% of the lands in Turkey are considered as forest areas. As it is known, forests are extremely important both for the country's economy and for a clean and sustainable ecosystem. Creating a tree inventory in our forests, which is a national wealth, is sustainable for the future. It is of vital importance in the development of afforestation policies and in preserving forest existence \cite{talay2019iha}. As a result of the tree inventory study, it is possible to extract information such as the number, characteristics, location, health status of existing tree species. Yilmaz et al. \cite{yilmaz2014insansiz}  as reported, the detection of information such as the boundaries of forest areas, the types and numbers of trees, height and location information allows many applications such as city planning, 3D city modeling, forestry and agricultural activities. Today, calculating the number of trees is a costly and error-prone process based on human observation and labor. Moreover, counting trees one by one by the staff poses a danger for some areas and carries the risk of making mistakes. On the other hand, sometimes using statistical methods, a generalization can be made in the selected geographical region based on the tree density in a certain area. This approach, in parallel with the convenience it provides, Calculations must be done carefully, as this may also cause an increase in the error rate. In order to overcome the problems mentioned above, it is possible to perform tree counting using image processing and machine learning. The basic requirements for this can be summarized as software that will calculate the number of trees as a result of processing the images obtained from the UAV. With the developing technology, the use of unmanned aerial vehicles (UAV) has become widespread. UAVs are used in many fields (military, emergency aid, agriculture, monitoring, security, etc.) within the framework of the license and authorization obtained from the general directorate of civil aviation. This study was carried out using DJI Advanced 4 Pro UAV. It is based on the principle of scanning the region at a certain height from the air and analyzing the captured images with the necessary software at the ground station. 

There are differences between UAV photogrammetry and conventional photogrammetry. UAV photogrammetry is designed for efficiency and using all data rather than accurate results in the context of local solution and local optimization. On the other hand, in traditional photogrammetry, global consistency, model validity, transactions are carried out within the framework of being correct, consistent, and compatible \cite{torun2017insansiz}. In order for UAV photogrammetry to replace classical photogrammetry, developments in two areas are required. The first is the mathematical/statistical model used in UAV photogrammetry and the adaptation of their application point to traditional photogrammetry. The second is that the sensor camera lens structure and lens distortion information, which forms the basis of the study of classical photogrammetry, and the possibilities provided by physical conditions for mathematical model design, can also be used for UAV sensors \cite{torun2016integrating, cryderman2014evaluation, draeyer2014white} . Although one of the best ways to obtain orthophoto maps is to use aircraft, some technical problems are encountered in this process. It is desirable that the images taken by airplanes or similar aircraft are in a vertical position relative to the earth as much as possible. However, in practice, this position is not possible due to the tilt and rotation of the aircraft. It is ideal if this curvature is usually less than 1°, and there are cases where it rarely exceeds 3°. Therefore, aerial photographs have such undesirable curvatures. Another problem with aerial photography is the so-called 'altitude shift'. This error consists of the height difference of artificial or natural objects in the field. Height shift points outward from picture midpoint is the slip. The height shift increases as you move away from the image midpoint. Apart from these errors, in aerial photographs; problems caused by film and photographic plane (film shrinkage, focal length error, etc.), distortions in camera lenses, distortions caused by atmospheric reflection and earth
problems such as distortions caused by its curvature may also occur. As a result, objects in the remotely sensed image may not be where they should be \cite{yilmaz2014insansiz,tao2004photogrammetric}. Moranduzzo and Melgani \cite{moranduzzo2013automatic} presented a SVM (Support Vector Machine) based solution to solve the problem of automobile detection and counting in images obtained by unmanned aerial vehicles (UAVs). Bazi et al. \cite{bazi2014automatic} developed an automated method for counting palm trees in UAV images. First, a number of key points are extracted using Scaled Invariant Feature Transformation. These key points were then analyzed with a pre-trained Extreme Learning Machine (ELM) classifier with and without a set of palm trees. As output, the ELM classifier marks each detected palm tree with a few key points. It is then proposed to combine these key points with an active contour method based on level sets to capture the shape of each tree. Finally, the texture of the regions obtained by the datasets was analyzed with local binary patterns to distinguish palm trees from other plants. Mohan et al. \cite{mohan2017individual} evaluated the applicability of low-consumer class cameras attached to the UAV and the applicability of the motion-structure algorithm for automatic individual tree detection using a local maxima-based algorithm in the Canopy Height Models obtained from the UAV. The number of trees was manually counted using the UAV-derived orthomosaic as a reference for each plot. A total of 367 reference trees were counted as part of the study and the algorithm detected 312 trees resulting in greater than 85\% accuracy (F-score 0.86). Overall, the algorithm missed 55 trees and incorrectly detected 46 trees resulting in a total tree count of 358. Shafri et al. \cite{shafri2011semi} utilizing high spatial resolution (1 m) airborne hyperspectral data, a diagram was created to use various approaches such as texture analysis, edge enhancement, morphology analysis, and bubble analysis to perform automatic tree counting. 

Santoso et al. \cite{santoso2016simple} developed a simple and user-friendly approach for the detection and counting of palm trees. As a result of the study, the researchers obtained an accuracy rate between 90\% and 95\%. Srestasathiern and Rakwatin \cite{srestasathiern2014oil} propose the local vertex detection hypothesis for the detection of trees. In this approach, each peak represents the highest point of each tree and the distinctive feature of the vegetation index is utilized. Using this approach, an accuracy rate of approximately 90\% has been achieved. Shao et al. \cite{shao1996model} carried out analyzes in the context of total biomass. According to the authors, it is of great importance to detect trees that are between certain areas for biological and commercial reasons. The determination of the number of trees in a particular location is an important indicator that reflects the productive capacity of the related species in that area. Site productive capacity refers to the total biomass (the amount of living beings in a given growing area) produced by a stand at a given site at all stages of its development, where the stand fully utilizes the resources necessary to grow wood. Avdan et al.\cite{avdan2014insansiz} conducted various researches on the field of archaeology by using the data obtained from unmanned aerial vehicles. In their study they produced orthophoto maps by changing the parameters of different heights and different overlay ratios. Muharrem and Murat \cite{ceylan3insansiz} created an orthophoto digital elevation model and a digital terrain model by processing the digital images they obtained using the UAV in Pix4D software. While making object-based classification, images were separated into meaningful clusters by segmentation and segments within the same threshold value were classified according to the determined index values. Gurbuz et al. \cite{gurbuz2016kentsel} performed tree detection from very high resolution RGB images obtained from UAV. For this, firstly, digital surface model (DEM) was obtained from the images by using orthophoto and automatic matching techniques. From the created orthophoto, four test areas were selected according to the tree density. segmentation and classification of the trees were made with object-based method in these test areas. In the next step, the peak points of the trees in the classified images were determined automatically. Finally, a reference data was created by obtaining the actual positions (geometric center points) of the trees by manual method, and accuracy analyzes were performed by comparing the peak points (local maximum) of the trees obtained by the automatic method with the reference data. 96\% , 82\%, 96\%, 47\% success rates was achieved in the first test, second, third and fourth test areas, respectively according to the results obtained.

%\paragraph
%The main objective of this study is to design a model that allows manipulation of only the desired attributes of an image. 
\section{Materials and Method}
\label{sec:TheMethod}

\subsection{UAV Image Acquisition}
\label{sec:Dataset_Pereparation}

\begin{figure}[!h]
	\centering
	\includegraphics[width=0.7\textwidth]{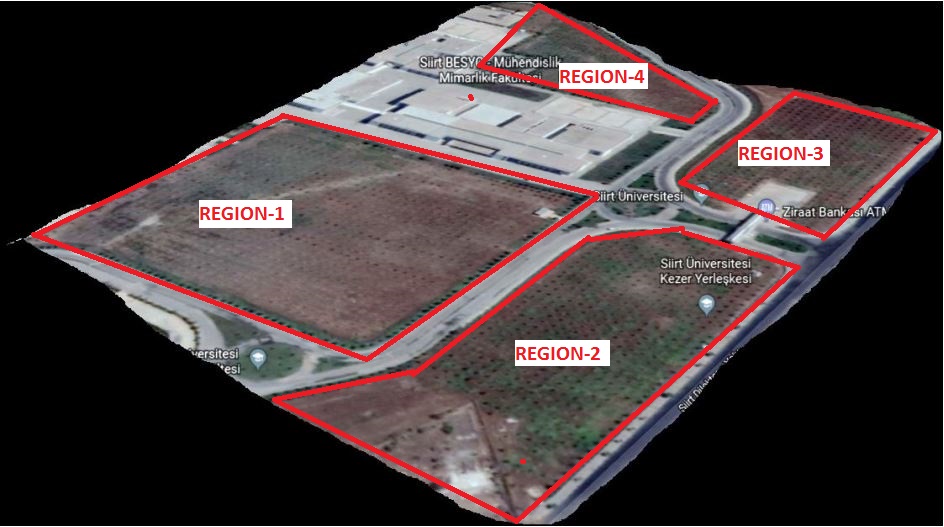}
	\caption{Designated four regions in the Kezer campus.}
	\label{fig:four_region}
\end{figure}

The geographical image of Siirt University Kezer Campus taken from the Google Earth application is shown in Fig. \ref{fig:four_region}. In terms of ease of access, four regions in Kezer Campus (Siirt University) were selected as the pilot application area, and then UAV scanned on these regions as shown in Fig. \ref{fig:four_region}. Coordinates of Region-1, Region-2, Region-3, and Region-4 are [37 57 47 N, 41 50 57 E], [37 57 38 N, 41 50 55 E], [37 57 44 N, 41 50 08 E], and [37 57 51 N, 41 51 06 E], respectively.  

\begin{figure}[!h]
	\centering
	\includegraphics[width=0.7\textwidth]{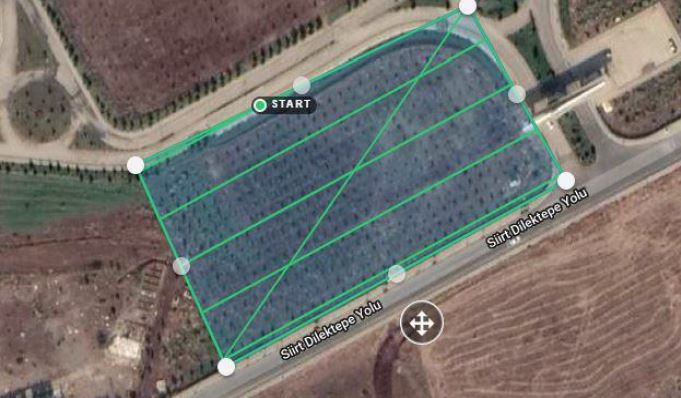}
	\caption{Typical path-ways determined via Drone-Deploy Software.}
	\label{fig:path_ways}
\end{figure}

250 color images were taken from various altitudes (30 m, 50 m, 75 m, and 90 m) with the UAV. These images were analyzed and it was determined that the ideal altitude with max gain in terms of image stitching was 50 meters. As depicted in Fig. \ref{fig:path_ways}, path-ways were determined with the Drone Deploy software for the area where the orthophoto map will be produced. Then, UAV flights were carried out over four regions. By default, the Drone Deploy application allows the UAV to take sequential images at certain time intervals during its flight over the paths determined in spatial regions. For image merging, we inferred that the ideal overlapping rate of consecutive images should be over 20\% as a result of exhaustive tests. Over 20\% overlapped images cause problems in the image assembly phase only. The high overlap rate means it takes longer for the UAV to scan the area. At the same time, the increase in the number of consecutive images causes the image merging process to take longer. However, this has been ignored in order for the image merging to work properly. 

\subsection{Dataset Pereparation}
\label{sec:Dataset_Pereparation}
\begin{figure}[!h]
	\centering
	\includegraphics[width=0.8\textwidth]{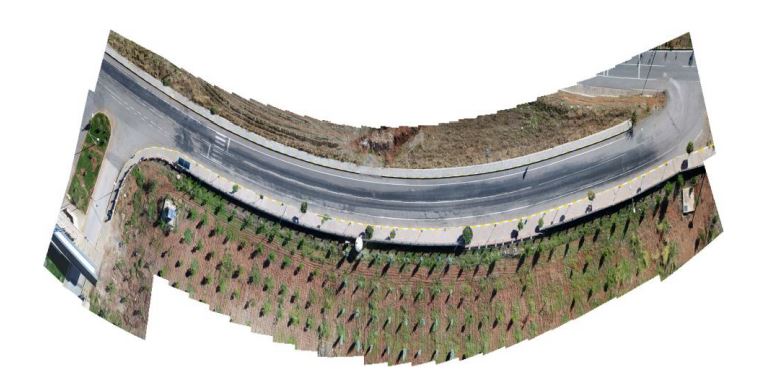}
	\caption{Prototypical stitched images taken from UAV.}
	\label{fig:stitched}
\end{figure}

The images obtained from the UAV were combined with Adobe Photoshop's image alignment method, and the scanned area was corrected to obtain an orthophoto map. Fig. \ref{fig:stitched} shows the combined representative scanned region as a result of photomerge from images with 30\% overlapped.The application we developed for the determination of the number of trees in our study basically selects according to the pixel density values of the images. In this context, the data set was prepared based on any image from the consecutive images taken by the UAV while the data set was being produced. Since the processing of the 20 MP image is not computationally cost efficient, the image to be used for the data set has been reduced to 1024x768 resolution. The Open Cezeri Library (OCL) developed for the Java ecosystem has been used in image processing operations \cite{atacs2016open}. OCL is basically a Domain Specific Language (DSL) framework developed in Siirt University Al-Jazari Cybernetics and Robotics laboratory, which allows the methods required for matrix, vectorization, image processing, machine learning and data visualization to be accessed through a single object. OCL acts on the principle of a static factory method with the CMatrix object on the basis of method chaining and Fluent Interface design pattern. The programmer does not need to know the remaining APIs that actually CMatrix calls in the background. He can call all the methods according to a certain flow (pipeline) logic and finish the application in a single line of code if he wants. You can access the developed OCL-based software codes from the Appendices section.

\begin{figure}[!h]
	\centering
	\includegraphics[width=0.8\textwidth]{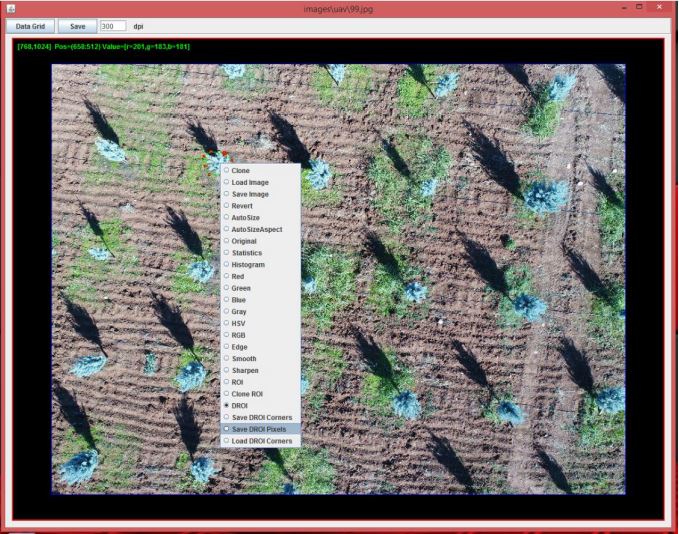}
	\caption{Dynamic ROI selection on the image for the data set generation.}
	\label{fig:droi}
\end{figure}

The high-resolution image is divided into grid-based thumbnail segments for cost-effective (memory-wise) processing. With a multi-threaded application developed, the obtained images were analyzed via various image processing techniques. As a result of the analysis, dynamic areas of interest DROI (Dynamic Region of Interest) were manually cropped from the image matrix to generate training and testing datasets for the machine learning algorithm based on the pixel density values and neighborhood in this DROI. In order to save the pixels belonging to the relevant class/group in the selection of the regions of interest, right-click on the image and select "DROI" (Dynamic Region of Interest) from the drop-down menu as shown in Fig. \ref{fig:droi}. In this study, DROI definitions were made for five different classes (i.e. tree-1, tree-2, shade, grass, and soil types) from images. After the desired region is marked with DROI, the coordinates of the pixels in the relevant region are saved in a text file. For this, "Save DROI pixels" must be selected from the drop-down menu. After this selection is performed, a sampling txt file for the relevant group/class is saved to the local disk. Then, N sample files saved for the relevant group are combined and converted into WEKA-ARFF file format, which allows machine learning to be trained by putting a class label on the last line.

\subsection{Classification}
\label{sec:Training_the_models}
%\paragraph
In this study, a pixel-value-based machine learning technique was used. For this, DROI regions were determined from different places on the image to be used in the train and test sets from the images obtained from the UAV, and the pixel values were utilized as the feature vector. In this context, three different modalities (RGB, HSV, and Gray) were determined and classification performances were evaluated. As the learning algorithms, Support Vector Machines (SVM), RepTree, Naive Bayes, and Multi-Layer Perceptron MLP from Artificial Neural Network models were used. These classifiers are the ones available by default in the Weka application. Since OCL can adopt the Weka library, classification algorithms are performed over OCL. Measurements of classification performances are also provided through the "FactoryEvaluation" class available in OCL. As mentioned before, the proposed tree number estimation model is based on pixels. To produce the learning dataset, DROI regions were determined from 10 different regions belonging to each class on a selected image and training and validation datasets were created for different modalities (RGB, HSV, GRAY). RGB, HSV and GRAY data sets consist of 71581 rows in total. 

\section{Results and Discussion}
\label{sec:experiments_and_results}
%\paragraph
Table 1 lists the 10-fold cross-validation performances of the classification algorithms of Naive Bayes, RepTree, SVM, MLP on different modalities.

\begin{table}[!h]
	\centering
	\caption{10-fold cross-validation performances of different classifiers in different modalities.}
	\label{tab:cross_validation}
	\begin{tabular}{lllll}
	\hline
	& \textbf{Classification} & \textbf{Accuracy} & \textbf{\% F-1 Score} \\ \hline
	\textbf{Model/Modalities} & \textbf{Naive Bayes} & \textbf{RepTree} & \textbf{SVM} & \textbf{MLP}\\ \hline
	GRAY & 48.19\%|0.455 & 49.54\%|0.488 & 47.67\%|0.488 & 45.53\%|0.431\\ \hline
	HSV & 85.33\%|0.851 & 85.44\%|0.854 & 74.56\%|0.741 & 87.91\%|0.880\\ \hline
	RGB & 59.85\%|0.597 & 86.19\%|0.862 & 88.47\%|0.885 & \textbf{88.89\%|0.889}\\ \hline
	\end{tabular}
\end{table}

When Table 1 is examined, it is seen that the best cross validation score can be reached with the MLP classifier in RGB format. Apart from that, it is also inferred that the GRAY format exhibits the lowest classification performance. The reason for this is that the difference between classes decreases in gray modality and is significantly mixed with each other. On the other hand, for all four classifiers, it is understood that the HSV format exhibits a classification performance close to that of RGB. Nevertheless, RGB has been chosen as the basic modality of the MLP classifier for use in the next section, since it performs better than HSV.Since the developed application will work in real-time, the machine learning algorithm should not go through the learning process every time, as in offline systems. Therefore, the MLP classifier is trained on the entire training dataset that we used for cross-validation in the previous step. The trained MLP classifier, which uses object serialization in JAVA, saves the connection weights of the trained model to the local disk as the "*.cls" extension. When performing real-time testing, serializing the relevant file into a trained MLP object saves a substantial time.

\begin{figure}[!h]
	\centering
	\includegraphics[width=0.6\textwidth]{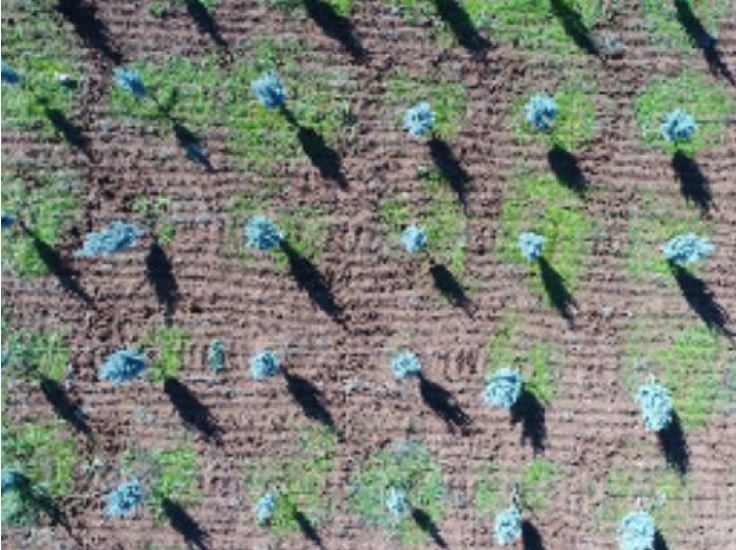}
	\caption{Original test image (approximately 24 trees can be counted in this image).}
	\label{fig:original_test_img}
\end{figure}

\begin{figure}[!h]
	\centering
	\includegraphics[width=0.6\textwidth]{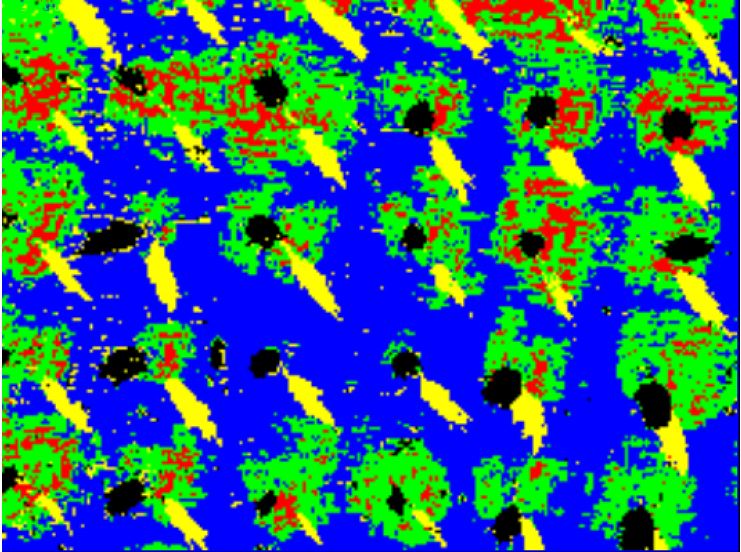}
	\caption{Pixel-wise segmented image after MLP classifier.}
	\label{fig:segmented_test_img}
\end{figure}

The previously trained MLP model subjects the original RGB image in Fig. \ref{fig:original_test_img} to a pixel-wise classification and creates a segmented image as in Fig. \ref{fig:segmented_test_img}.

\begin{figure}[!h]
	\centering
	\includegraphics[width=0.6\textwidth]{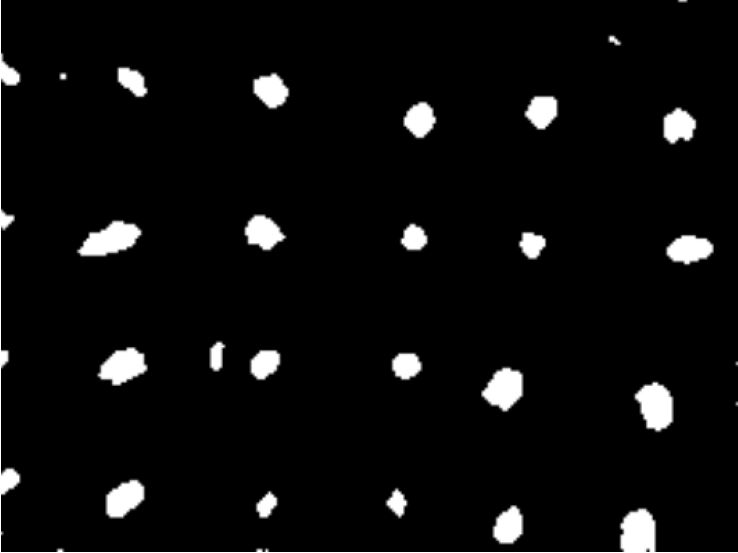}
	\caption{Filtered image in terms of black pixels (tree-1 group) as a binary format.}
	\label{fig:filtered_img}
\end{figure}

Here, black colors represent tree-1 class, blue color soil class, yellow color shades, green color grass and red spots frequently seen in green areas represent tree-2 class. After this stage, in order to determine the number of the tree-1 class, only the black pixels are filtered from the color image as shown in Fig. \ref{fig:filtered_img}. Then, the obtained regions (blobs) are determined and counting is attempted.

\begin{figure}[!h]
	\centering
	\includegraphics[width=0.6\textwidth]{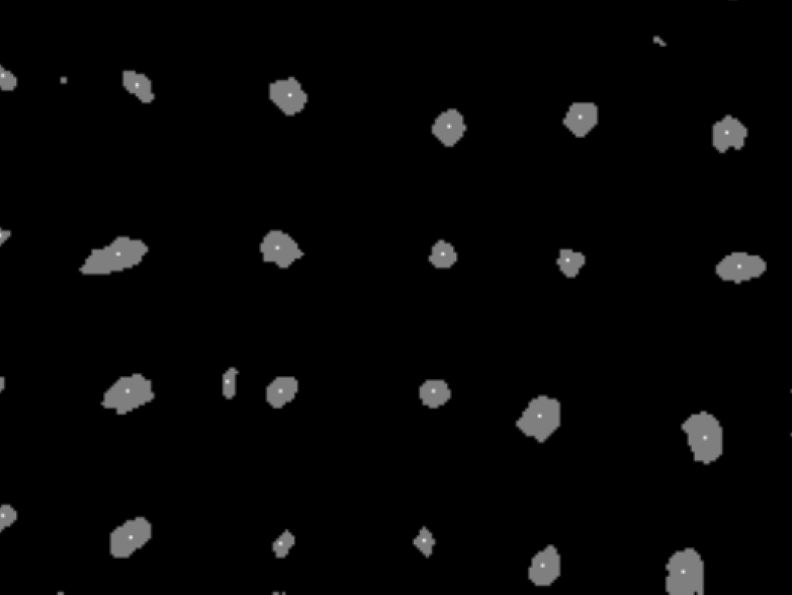}
	\caption{Detection the centroid of the blobs.}
	\label{fig:centroid_img}
\end{figure}

When the image in Fig. \ref{fig:filtered_img} is examined, it will be seen that there are some small blobs that the median filter cannot destroy, but that are not assigned to the tree class. It has been seen that the blob detection algorithm we have developed contributes to the detection rate by disabling blobs whose size is below a certain value. In summary, the blob detection algorithm works as follows. The binary image is scanned line by line from the top left position in the form of a matrix (there are only two values in the matrix, the pixel value of the ground is 0, and the pixel values of the blobs are set as 255). When the algorithm finds the value of 255 from the pixels in the line it scans (by going column-by-column to the right), it assigns it as the start of a new blob. It looks at the neighborhood status in the next pixels and if there is 8 neighborhood status (right, left, bottom, top, and diagonal), the relevant pixel is assigned to the most suitable blob. After it is assigned, the pixel density value is reduced from 255 (white) to 127 (gray) so that it is not checked again. After all these processes are finished, the centroids of all blobs in the blob list are found. Fig. \ref{fig:centroid_img} shows the detected blobs and their centers of gravity.

\begin{figure}[!h]
	\centering
	\includegraphics[width=0.6\textwidth]{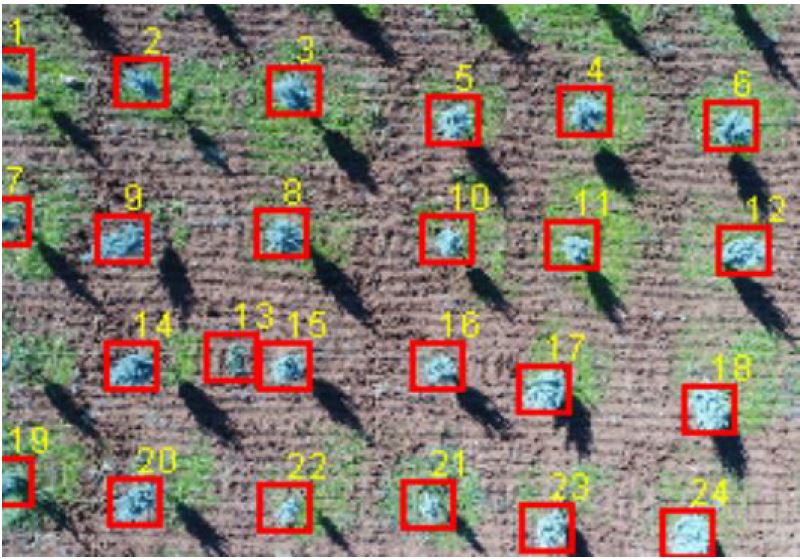}
	\caption{Detected and counted tree objects on the test image depicted as yellow bounding box.}
	\label{fig:bounding_img}
\end{figure}

Next, it is time to mark the blobs whose centroids have been determined on the original image. For this, with the help of the drawRect and drawString methods defined in OCL, overlay is performed on the original color picture. Fig. \ref{fig:bounding_img} shows the tree regions (red rectangle) and tree count number (top yellow) superimposed/drawn on the original test image.

Table 2 lists the accuracy rates in the regions estimated using the MLP classifier.

\begin{table}[!h]
	\centering
	\caption{ Prediction performance of MLP classifier.}
	\label{tab:test_perf}
	\begin{tabular}{lllll}
	\hline
	\textbf{Regions} & & \textbf{Prediction} & \textbf{Performance}  \\ \hline
	Region-1 & & & \%89 & \\
	Region-2 & & & \%91 & \\
	Region-3 & & & \%75 & \\
	Region-4 & & & \%93 & \\ \hline

	\end{tabular}
\end{table}

\section{Conclusion and Future Works}
\label{sec:conclusion_and_future_works}
%\paragraph

In this study, aerial images were obtained using the DJI Phantom 4 Advanced drone. Tree detection and counting processes were carried out in 4 different regions in the Kezer campus of Siirt University. With the OCL software framework, it has been tried to determine the tree numbers on the image by pixel-based filtering and blob detection without using any commercial or open source applications. MLP classifier and RGB modality with 10-fold cross validation performance were chosen as reference and used for tree count estimation. The application developed with an average accuracy rate of 87\% is considered successful. It is thought that the software we have developed can be used to calculate the number of other objects such as vehicles, people, buildings in future projects.

\section*{Acknowledgements}
This study was funded by Siirt University Scientific Research Projects Grant (SIÜBAP) within the scope of project numbered 2017-SIÜFEB-21.

\bibliographystyle{ieeetr}
%\footnotesize \bibliography{reference}

\begin{thebibliography}{10}

\bibitem{fao}
FAO, ``{State of the World's Forests}.''
  \url{https://www.fao.org/state-of-forests/en/}, 2019.
\newblock [Online; accessed 19-June-2021].

\bibitem{talay2019iha}
A.~Talay, ``{\.I}ha imgelerinden bilgisayar g{\"o}r{\"u}s{\"u} kullan{\i}larak
  a{\u{g}}a{\c{c}} say{\i}s{\i} kestirimi,'' Master's thesis, Fen Bilimleri
  Enstit{\"u}s{\"u}, 2019.

\bibitem{yilmaz2014insansiz}
V.~Y{\i}lmaz, A.~Akar, {\"O}.~Akar, O.~G{\"u}ng{\"o}r, F.~Karsl{\i}, and
  E.~G{\"o}kalp, ``{\.I}nsansiz hava araci {\.i}le {\"u}ret{\.i}len ortofoto
  har{\.i}talarda do{\u{g}}ruluk anal{\.i}z{\.i},'' {\em Uzaktan Alg{\i}lama ve
  Co{\u{g}}rafi Bilgi Sistemleri Sempozyumu (UZAL-CBS 2014)}, pp.~14--17, 2014.

\bibitem{torun2017insansiz}
A.~Torun, ``{\.I}nsans{\i}z hava arac{\i} ({\.i}ha) sekt{\"o}r{\"u}nde trend:
  {\.I}ha fotogrametrisi bak{\i}{\c{s}}{\i}yla,'' {\em Afyon Kocatepe
  {\"U}niversitesi Fen Ve M{\"u}hendislik Bilimleri Dergisi}, vol.~17, no.~4,
  pp.~35--52, 2017.

\bibitem{torun2016integrating}
A.~Torun, ``Integrating geospatial technologies: Reflections on intergeo
  2016,'' {\em GIM International, November}, 2016.

\bibitem{cryderman2014evaluation}
C.~Cryderman, S.~B. Mah, and A.~Shufletoski, ``Evaluation of uav
  photogrammetric accuracy for mapping and earthworks computations,'' {\em
  Geomatica}, vol.~68, no.~4, pp.~309--317, 2014.

\bibitem{draeyer2014white}
B.~Draeyer and C.~Strecha, ``White paper: How accurate are uav surveying
  methods,'' {\em Pix4D White Paper}, 2014.

\bibitem{tao2004photogrammetric}
C.~V. Tao, Y.~Hu, and W.~Jiang, ``Photogrammetric exploitation of ikonos
  imagery for mapping applications,'' {\em International Journal of Remote
  Sensing}, vol.~25, no.~14, pp.~2833--2853, 2004.

\bibitem{moranduzzo2013automatic}
T.~Moranduzzo and F.~Melgani, ``Automatic car counting method for unmanned
  aerial vehicle images,'' {\em IEEE Transactions on Geoscience and Remote
  Sensing}, vol.~52, no.~3, pp.~1635--1647, 2013.

\bibitem{bazi2014automatic}
Y.~Bazi, S.~Malek, N.~Alajlan, and H.~AlHichri, ``An automatic approach for
  palm tree counting in uav images,'' in {\em 2014 IEEE Geoscience and Remote
  Sensing Symposium}, pp.~537--540, IEEE, 2014.

\bibitem{mohan2017individual}
M.~Mohan, C.~A. Silva, C.~Klauberg, P.~Jat, G.~Catts, A.~Cardil, A.~T. Hudak,
  and M.~Dia, ``Individual tree detection from unmanned aerial vehicle (uav)
  derived canopy height model in an open canopy mixed conifer forest,'' {\em
  Forests}, vol.~8, no.~9, p.~340, 2017.

\bibitem{shafri2011semi}
H.~Z. Shafri, N.~Hamdan, and M.~I. Saripan, ``Semi-automatic detection and
  counting of oil palm trees from high spatial resolution airborne imagery,''
  {\em International journal of remote sensing}, vol.~32, no.~8,
  pp.~2095--2115, 2011.

\bibitem{santoso2016simple}
H.~Santoso, H.~Tani, and X.~Wang, ``A simple method for detection and counting
  of oil palm trees using high-resolution multispectral satellite imagery,''
  {\em International journal of remote sensing}, vol.~37, no.~21,
  pp.~5122--5134, 2016.

\bibitem{srestasathiern2014oil}
P.~Srestasathiern and P.~Rakwatin, ``Oil palm tree detection with high
  resolution multi-spectral satellite imagery,'' {\em Remote Sensing}, vol.~6,
  no.~10, pp.~9749--9774, 2014.

\bibitem{shao1996model}
Y.~Shao, M.~R. Raupach, and J.~F. Leys, ``A model for predicting aeolian sand
  drift and dust entrainment on scales from paddock to region,'' {\em Soil
  Research}, vol.~34, no.~3, pp.~309--342, 1996.

\bibitem{avdan2014insansiz}
U.~Avdan, E.~{\c{S}}enkal, R.~{\c{C}}{\"o}mert, and S.~Tuncer,
  ``{\.I}nsans{\i}z hava arac{\i} ile olu{\c{s}}turulan verilerin
  do{\u{g}}ruluk analizi,'' {\em 5. Uzaktan Alg{\i}lama ve Co{\u{g}}rafi Bilgi
  Sistemleri Sempozyumu (UZAL-CBS 2014)}, pp.~14--17, 2014.

\bibitem{ceylan3insansiz}
M.~C. Ceylan and M.~Uysal, ``{\.I}nsans{\i}z hava arac{\i} ile elde edilen
  veriler yard{\i}m{\i}yla a{\u{g}}a{\c{c}} {\c{c}}{\i}kar{\i}m{\i},'' {\em
  T{\"u}rkiye Fotogrametri Dergisi}, vol.~3, no.~1, pp.~15--21.

\bibitem{gurbuz2016kentsel}
M.~G{\"u}rb{\"u}z, {\em Kentsel Alanlarda {\.I}ha G{\"o}r{\"u}nt{\"u}lerinden
  Ortofoto Olu{\c{s}}turma Ve Otomatik A{\u{g}}a{\c{c}} Tespiti}.
\newblock PhD thesis, Y{\"u}ksek Lisans Tezi, Hacettepe {\"U}niversitesi, Fen
  Bilimleri Enstit{\"u}s{\"u}, 2016.

\bibitem{atacs2016open}
M.~Ata{\c{s}}, ``Open cezeri library: A novel java based matrix and computer
  vision framework,'' {\em Computer Applications in Engineering Education},
  vol.~24, no.~5, pp.~736--743, 2016.

\end{thebibliography}

\end{document}